\documentclass{article} 
\usepackage{iclr2024_conference,times}
\usepackage{algorithm2e}

\usepackage{amsmath,amsfonts,bm}

\def\eqref#1{equation~\ref{#1}}

\def\1{\bm{1}}

\DeclareMathAlphabet{\mathsfit}{\encodingdefault}{\sfdefault}{m}{sl}
\SetMathAlphabet{\mathsfit}{bold}{\encodingdefault}{\sfdefault}{bx}{n}

\usepackage{hyperref}
\usepackage{url}
\usepackage{cleveref}
\usepackage[pdftex]{graphicx}
\usepackage{booktabs} 
\usepackage{multirow}
\usepackage[inline]{enumitem}

\usepackage{caption}
\usepackage{subcaption}
\usepackage[normalem]{ulem}
\usepackage[
    acronym,
]{glossaries}
\usepackage{cleveref}
\usepackage{csquotes}

\newacronym{fl}{FL}{Federated Learning}
\newacronym{clip}{CLIP}{Contrastive Language-Image Pretraining}
\newacronym{fedtpg}{FedTPG}{\textbf{Fed}erated \textbf{T}ext-driven \textbf{P}rompt \textbf{G}eneration}
\newacronym{coop}{CoOp}{Context Optimization}
\newacronym{fedcoop}{FedCoOp}{Federated Context Optimization}
\newacronym{fedkgcoop}{FedKgCoOp}{Federated Knowledge-guided Context Optimization}

\newif\ifdraft

\drafttrue

\title{Text-driven Prompt Generation for Vision-Language
Models in Federated Learning}

\author{%
  Chen Qiu\thanks{Equal contribution. \quad Correspondence to: Chen.Qiu@us.bosch.com}\\
  Bosch Center for AI, USA \\
  \And
  Xingyu Li$^{*}$\thanks{Work done during internship at Bosch Center for AI, USA.}\\
  Tulane University\\
  \And
 Chaithanya Kumar Mummadi \& Madan Ravi Ganesh \& Zhenzhen Li\\
 Bosch Center for AI, USA\\
 \And
 Lu Peng\\ Tulane University\\
 \And
 Wan-Yi Lin\\ Bosch Center for AI, USA\\
}

\definecolor{mybrown}{RGB}{165, 42, 42}

\iclrfinalcopy 
\begin{document}

\maketitle

\begin{abstract}
    Prompt learning for vision-language models, e.g., CoOp, has shown great success in adapting CLIP to different downstream tasks, making it a promising solution for federated learning due to computational reasons. Existing prompt learning techniques replace hand-crafted text prompts with learned vectors that offer improvements on seen classes, but struggle to generalize to unseen classes. Our work addresses this challenge by proposing Federated Text-driven Prompt Generation (FedTPG), which learns a unified prompt generation network across multiple remote clients in a scalable manner. The prompt generation network is conditioned on task-related text input, thus is context-aware, making it suitable to generalize for both seen and unseen classes. Our comprehensive empirical evaluations on nine diverse image classification datasets show that our method is superior to existing federated prompt learning methods, that achieve overall better generalization on both seen and unseen classes and is also generalizable to unseen datasets.

\end{abstract}

\section{Introduction}
Vision-language models have recently emerged as a transformative technology for machine learning applications. Seminal contributions like \gls*{clip} \cite{radford21a} have demonstrated unprecedented capabilities in diverse image classification tasks. Different classification methods often leverage manually-engineered text prompts, such as ``a photo of a [class]," to utilize \gls*{clip}'s rich semantic features \citep{jia2021scaling}.
\gls*{clip} has shown its robustness and versatility in handling a wide range of image distributions. These properties make \gls*{clip} naturally aligned with the objective of \gls*{fl}, 
a decentralized approach to train machine learning models with data privacy. However,  high computational and communication costs associated with server-client interaction make the training of \gls*{clip} impractical in the \gls*{fl} setting. This motivates us to explore more efficient and effective methods to adapt the advantages of \gls*{clip} in \gls*{fl}. 

Emerging prompt learning methodologies based on \gls*{clip} such as \gls*{coop} have revealed that fine-tuning \gls*{clip} can be made more efficient by substituting hand-crafted prompts with learnable soft prompt vectors in a few-shot learning paradigm \citep{perez2021true} for one downstream task in centralized learning \citep{zhou2022coop,zhou2022cocoop,zhu2022prompt,Yao_2023_CVPR}.
Existing federated prompt learning method, \gls*{fedcoop} \citep{guo2023promptfl}, adapts the learning paradigm of \gls*{coop} to \gls*{fl} by learning a unified set of prompt vectors across multiple clients with different datasets. \gls*{fedcoop} improves over \gls*{clip} on the seen (during training) classes in each client, but it struggles to generalize on the unseen classes (not included in training).
Similarly, prompt vectors optimized on seen classification tasks fail to generalize to new tasks of different contexts (e.g., from object recognition to texture classification). 
Unless otherwise noted, we refer to \enquote{task} as an image classification dataset within the context of this work. 

\begin{figure}[t]
    \vspace{-10pt}
        \centering
            \includegraphics[width=0.88\linewidth]{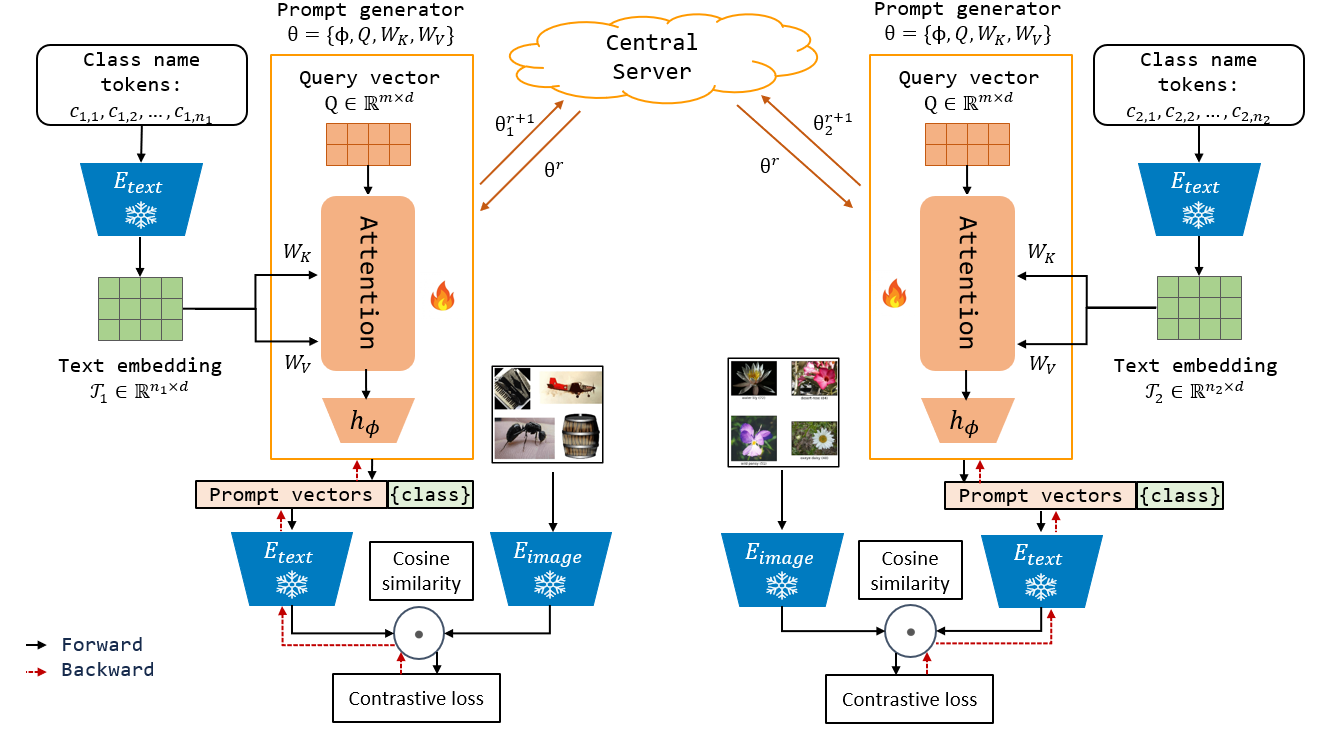}
        \caption{Our proposed \gls*{fedtpg} learns a unified prompt generator over the frozen \gls*{clip} model for converting task-related text input $\mathcal{T}$ to context-aware prompt vectors. The prompt generator is learned across multiple clients with different classification datasets collaboratively.}
        \label{fig:system}
    \vspace{-10pt}
    \end{figure}

Instead of learning one unified set of prompt vectors for different classification tasks, we propose to convert text input containing task-specific semantic information to context-aware prompt vectors.
Benefiting from context information in text input, 
we aim to generate prompt vectors that generalize well to classification tasks that have not been previously observed (refer \Cref{fig:tpg} for an illustration of the concept).
Following that, we propose \gls*{fedtpg}, which learns a lightweight unified prompt generator across multiple clients collaboratively. 
Each client optimizes the prompt generator locally for its classification task described by few-shot image-text pairs, followed by the FL server-client communication to obtain the global prompt generator model. An overview of our \gls*{fedtpg} with two remote clients is shown in \Cref{fig:system}.
By training on various image classification tasks, our prompt generator learns to generate prompt vectors conditioned on context-related text inputs. Leveraging contextual awareness, the generated prompt vectors differentiate themselves across various tasks and enrich \gls*{clip} with context information of the target task. 
Our comprehensive evaluation on nine diverse image classification datasets demonstrate that \gls*{fedtpg} has improved generalization over the existing prompt learning method \gls*{fedcoop} on unseen classes 
by $4.32\%$ and unseen datasets by $1.82\%$, on average.

We summarize the contributions of our work as follows: (1) We develop a text-driven prompt generation (TPG) technique to improve the generalization performance from observed classification tasks to new classification tasks with different contexts. Instead of learning fixed prompt vectors, the prompt generator converts task-related text input to context-aware prompt vectors for various image classification tasks. (2) We propose \gls*{fedtpg}, a scalable way of learning a unified, generalized text-driven prompt generator  across multiple clients with various classification tasks collaboratively. (3) We undertake exhaustive empirical analysis using nine datasets to validate the efficacy of \gls*{fedtpg}. Our comparative studies with existing federated prompt learning methods demonstrate \gls*{fedtpg}'s superior generalization performance on image classification tasks encompassing a range of domains.

\section{Related Work}

\textbf{Visual-Language Model Prompt Learning.}
Prompt learning, a variation of fine-tuning Vision-Language Models (VLMs), has shown considerable promise in enhancing the task-specific performance of existing pre-trained models under few-shot settings. 
A significant advancement in this direction was CoOp~\citep{zhou2022coop}, which introduced the notion of optimizing continual prompt context vectors for better task adaptation. 
CoCoOp~\citep{zhou2022cocoop} further improves CoOp by combining an input image-conditioned token with the learnable prompt context vectors through the use of a lightweight neural network. 
Several other works have also explored the interplay between textual prompts and visual inputs~\citep{zang2022unified, li2023efficient}.
Specifically, MaPLe~\citep{Khattak_2023_CVPR} extends the prompt learning paradigm to multi-modal tasks, integrating both visual and textual information for a more robust task adaptation.
On the other hand, LASP~\citep{Bulat_2023_CVPR} and KgCoOp~\citep{Yao_2023_CVPR} explore text-to-text optimization to encourage human language-aware soft prompting in VLMs. 
In this paper, we focus on improving the prompt learning, making it generalizes well to unseen tasks with the context-aware text input information.




\textbf{Federated Learning with Visual-Language Models.}
Federated Learning (FL)~\citep{mcmahan2017communication} has emerged as a pivotal paradigm for decentralized training of machine learning models on heterogeneous data~\citep{li2023fedlga}, thereby preserving data privacy~\citep{li2021lomar} and reducing data transfer overhead~\citep{qu2022convergence}. 
Recently, fine-tuning of VLMs has been extended to the federated setting to reduce the computational burden on a single device while addressing existing issues in FL like poor performance and robustness under cross-domain settings, non-IID data distribution across clients, and others.
FedCLIP~\citep{lu2023fedclip} proposes a direct extension of standard fine-tuning of CLIP to the federated learning setting to enable strong performance and personalization. 
From a data heterogeneity perspective, \cite{halbe2023hepco} provides a continual lifelong prompt learning mechanism to mitigate the effect of client drift. 
\cite{wang2023cooperative} further showcase the corrective attribute of prompts when trained under hardware-aware settings in the snapshot compressive imaging application domain while \cite{chen2023spatial} highlight the adaptability of federated prompt-based methods to diverse data landscapes beyond visual and textual data, in their case for weather forecasting.
Of relevance to our approach is PromptFL~\citep{guo2023promptfl}\footnote{For the sake of presentation, we name PromptFL as FedCoOp, as PromptFL adapts CoOp to the FL setting.}, which proposes a federated learning framework for prompt learning that enables participants to cooperatively learn a common prompt vector.  \cite{su2022cross} who delve into the cross-domain applicability of federated prompt learning in VLMs, and \cite{guo2023pfedprompt} who combine a federated prompt learning scheme with personalized spacial visual features.
A key distinction between these methods and our approach to federated prompt learning is our use of a learnable text-conditioned prompt generator which improves generalization performance on both seen and unseen tasks, a typically unexplored setting for VLMs under the FL scheme.
Concurrent with our work, \cite{yang2023efficient} propose a prompt generator with a cross-attention mechanism similar to our approach. 
In addition to their focus on using a frozen ViT backend, 
we hypothesize that their dependence on fixed client-specific features learned for seen clients would limit their generalization to unseen tasks. 
In comparison, our prompt generation depending on text inputs has no hurdles in generalizing to unseen tasks.

\section{Method}
\RestyleAlgo{ruled}
\SetKwInput{kwData}{Data}
\SetKwInput{kwModel}{Model}
\SetKwInput{kwInput}{Input}
\SetKwInput{kwOutput}{Output}
In this section, we present our problem setup of federated learning 
in~\Cref{sec:statement}, followed by our text-driven prompt generation technique in~\Cref{sec:prompt_generation} and finally propose our \gls*{fedtpg} algorithm that deploys text-driven prompt generation in federated learning in~\Cref{sec:federated_learning}.

\subsection{Problem setup}
 \label{sec:statement}

We consider a federated network setting with one central server for model aggregation and multiple remote clients, 
where each client $i$ has a private classification dataset with labeled images $(x,y)\sim \mathcal{D}_i$ from $n_i$ classes with class name tokens $\{c_{i,j}\}_{j=1}^{n_i}$ (a sample setting with two remote clients is depicted in~\Cref{fig:system}). Data distribution across the federated network follows a non-IID setup where clients contain samples from a disjoint set of classes.
The goal of \gls*{fl} framework in our setup is to jointly learn one model that not only solves different image classification tasks spanning across multiple remote clients but also attains generalization ability to unseen classes and datasets. 
In contrast to the setting in \gls*{fl} literature \citep{kairouz2021advances}, our consideration of generalization to unseen classes and datasets makes our setup more challenging.
Following the recent success of vision-language models like \gls*{clip} across a broad range of tasks \citep{radford21a}, we look into the adaptation of \gls*{clip} models in our \gls*{fl} framework to achieve our goal of generalization.
\gls*{clip} is a large vision-language model with an image encoder $E_{image}$ and a text encoder $E_{text}$, and can classify images utilizing linguistic knowledge. In our \gls*{fl} setup, we consider each client to have access to an off-the-shelf publicly available pretrained \gls*{clip} model. Here, we focus on adapting the pretrained \gls*{clip} model collaboratively across all clients. However, updating large models like \gls*{clip} across numerous remote clients requires extensive computational power and bandwidth, making it impractical for \gls*{fl} applications. Recently, prompt learning has been used to offer a computation and communication efficient federated learning framework e.g., \gls*{fedcoop} \citep{guo2023promptfl} for adapting a frozen \gls*{clip} across multiple clients. Specifically, hand-crafted text prompts (e.g., ``a photo of a [class]'') for $E_{text}$ are replaced with trainable soft prompt vectors $v_1,v_2,...,v_m$, while keeping CLIP weights unaltered. In federated prompt learning, lightweight trainable prompt vectors are shared across clients at each communication round and updated with local training on client data.

In this work, our goal is to learn a \gls*{fl} prompt model that can solve various image classification tasks across multiple clients and also generalize to novel classes or image classification tasks from new clients, which can be challenging to existing methods like \gls*{fedcoop}. \citet{zhou2022cocoop} have shown that \gls*{coop}'s prompt vectors, optimized for observed classes, fail to generalize to novel classes. We notice a similar generalization issue in \gls*{fedcoop} i.e., learned unified prompt vectors perform well on the seen classification tasks across remote clients, but fail to generalize to tasks with different contexts (e.g., from object recognition to texture classification). 
We attribute this behavior to the fixed nature of soft prompts and not being able to adjust to the context of the task. 
To address this, we propose a novel strategy that alters how the soft prompt vectors are obtained. Instead of directly learning the soft prompts, we learn a text-driven prompt generation 
module that takes task-related text input and transforms it into context-aware prompt vectors, which we detail in the next section. 
\begin{figure}[t]
    \vspace{-15pt}
        \centering
            \includegraphics[width=0.82\linewidth]{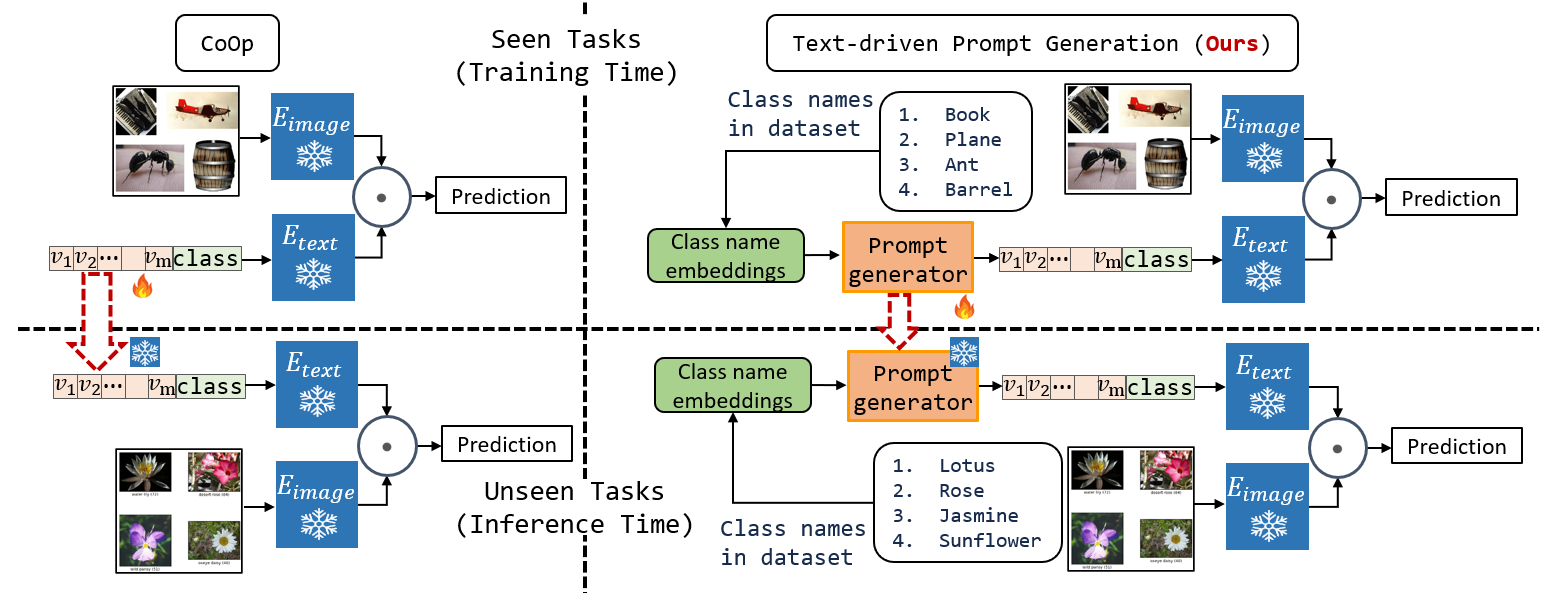}
        \caption{Our proposed prompt generator generates prompt vectors conditioning on the targeted classification task-related text input. Leveraging with contextual awareness, the generated prompt vectors enrich CLIP with context information in the text input and can generalize to unseen classes.}
        \label{fig:tpg}
    \vspace{-10pt}
    \end{figure}
\subsection{Text-driven Prompt Generation}\label{sec:prompt_generation}
We develop a prompt generation module $f_\theta$ that generates context-aware prompt vectors conditioned on the target classification task-related text inputs, as shown in \Cref{fig:tpg}. The text input is translated to text embeddings $\mathcal{T}$ and the prompt generator $f_\theta$ converts these text embeddings $\mathcal{T}$ to a set of $m$-length input prompt vectors 
$\mathcal{P} \in\mathbb{R}^{m\times d}$ 
for $E_{text}$
as: 
\begin{align}\label{eqn:prompt_generate}
    \mathcal{P} = \{v_k\}_{k=1}^m = f_\theta(\mathcal{T}).
\end{align}
Context-related text input can be obtained from the natural language description.
We find that available candidate class names naturally represent context-related text for the classification task ([class $0$], [class $1$], ..., [class $n$]).
We translate the natural language class names to text embeddings as $\mathcal{T} = \{E_{text}(c_j)\}_{j=1}^n$ $\in\mathbb{R}^{n\times d}$, 
a set of embeddings of $n$ class name tokens $c_j$ from \gls*{clip} text encoder\footnote{For simplicity, we consider a single client with index $i=1$, and remove the client's index $i$ in notations.}.
Besides, prompt generator $f_\theta$ is a lightweight cross-attention module comprising of learnable parameters $\phi, Q\in\mathbb{R}^{m\times d}, W_K\in\mathbb{R}^{d\times d}, W_V\in\mathbb{R}^{d\times d}$. Given the text embeddings $\mathcal{T}$ we have:
\begin{align}
    f_\theta(\mathcal{T}) = h_\phi(\text{CrossAttention}(Q, K_\mathcal{T}, V_\mathcal{T})) 
    \quad \text{with} \quad K_\mathcal{T} = \mathcal{T} \times W_K, \quad V_\mathcal{T} = \mathcal{T} \times W_V.
\end{align}
The prompt generator transforms context information from the text embeddings $\mathcal{T}$ into key and value vectors $K_\mathcal{T}$ and $V_\mathcal{T}$ respectively.
Cross-attention layer merges these vectors with the learnable query vector $Q$, 
and hidden layers $h_\phi$ projects cross-attention layer output to prompt vectors $\mathcal{P}$.

Prompt vector for each class $j$ is defined as $t_j= \mathcal{P} \cup \{c_j\}$, concatenating generated context prompt vectors $\mathcal{P}$ and text token of class name $c_j$.
Given an input image $x$ and prompt vectors for all $n$ candidate classes, the prediction probability of \gls*{clip} for a classification task is computed as follows:
\begin{align}
\label{eqn:fedtpg_pred}
p_{\theta}(y = j|x,\mathcal{T}) &= \frac{\exp(\cos(E_{image}(x),E_{text}(t_j))/\tau)}{\sum_i^n\exp(\cos(E_{image}(x),E_{text}(t_i))/\tau)}. 
\end{align}
Text embeddings $\mathcal{T}$ produced from a well-pretrained text encoder like CLIP provides rich and meaningful context information for a given text.
The prompt generator $f_\theta$ should serve to extract and transfer context-critical information from the already meaningful embeddings $\mathcal{T}$ to prompt vectors $\mathcal{P}$. 
Training $f_\theta$ on different classification tasks from diverse contexts would facilitate its convergence to produce generalized context-aware prompt vectors, and thus improve prediction precision of $p_{\theta}(y = j|x,\mathcal{T})$ on unseen classes. 
In practical scenarios, the data encompassing a wide range of classification tasks is typically distributed across different clients. Addressing this, we next present a scalable way of learning the prompt generator across multiple clients collaboratively.

\begin{algorithm}[t!]
    \caption{\gls*{fedtpg} Algorithm}\label{algo:fedtpg}
    \kwInput{No. of communication rounds $R$, No. of local epochs $K$, initialization parameters $\theta^{0}$. } 
    \textbf{Server executes:}\\
    Initialize prompt generator $f_\theta$ parameters with $\theta^{0}$. \\
    \For{$r\gets0$ \KwTo $R$}{
        Pick a random subset of remote clients as $\mathcal{S}^r$. \\
        \For{$i\in \mathcal{S}^r$ \emph{in parallel}}{
            Send the current global model $\theta^{r}$ to client $i$.\\
            Receive locally updated $\theta^{r+1}_{i}$ from \textbf{Local Client Training}.
        }
        Aggregate the updated model parameters $\theta^{r+1} = \frac{1}{|\mathcal{S}^r|} \sum_{i \in \mathcal{S}^r} \theta^{r+1}_{i}$.\\
        }
    Obtain the final model parameter $\theta^{R}$.\\
    \textbf{Local Client Training:}\\
    Obtain the set of class name embeddings $\mathcal{T}_i = \{E_{text}(c_{i,j})\}_{j=1}^{n_i}$.\\
    \For{$k\gets0$ \KwTo $K$}{
        Generate the context prompt vectors $\mathcal{P}^{r}_{i} = f_{\theta_{i}^{r}}(\mathcal{T}_i)$.\\
        Get the prompt vectors for each class 
        $t^r_{i,j} = \mathcal{P}^{r}_{i} \cup \{c_{i,j}\}$.\\
        Update parameters $\theta^{r}$ to $\theta_{i}^{r+1}$ locally using \cref{eqn:fedtpg_pred,eqn:clip_loss,eqn:sgdk} on $(x, y)\sim \mathcal{D}_i$.\\ }
\end{algorithm}

\subsection{Local Training and Server Aggregation}
\label{sec:federated_learning}
We incorporate our prompt generation module in \gls*{fl} settings, where multiple remote clients handling diverse image classification tasks train the prompt generator $f_\theta$ collaboratively. We refer this approach as \textbf{Fed}erated \textbf{T}ext-driven \textbf{P}rompt \textbf{G}eneration (\gls*{fedtpg}).
We outline the training pipeline of our \gls*{fedtpg} in Algorithm~\ref{algo:fedtpg}. 
Initially, the server initializes $f_\theta$ parameters randomly with $\theta^{0}$ and then at each communication round, a random subset of remote clients $\mathcal{S}^r$ retrieve the up-to-date $f_\theta$ parameters for local training. Below we describe the training steps of \gls*{fedtpg} at each round $r$: 
\begin{itemize}[leftmargin=*]
    \item \textbf{Step I:} Remote client $i$ in $\mathcal{S}^r$ receives current up-to-date parameters $\theta^{r}$ to configure the local $f_{\theta^{r}}$.
    \item \textbf{Step II:} At each client, the frozen \gls*{clip} text encoder $E_{text}$ provides text embeddings of the local available class name tokens $\mathcal{T}_i = \{E_{text}(c_{i,j})\}_{j=1}^{n_i}$. The prompt generator $f_{\theta^{r}}$, the frozen \gls*{clip} model, the context text embeddings $\mathcal{T}_i$, and the dataset $\mathcal{D}_i$ together define the local objective as: 
    \begin{align}\label{eqn:clip_loss}
        L_i(\theta^r; \mathcal{T}_i) = -\mathbb{E}_{(x,y)\in \mathcal{D}_i}y\log p_{\theta^r}(y|x,\mathcal{T}_i),
     \end{align}
     where $p_{\theta^r}(y|x,\mathcal{T}_i)$ is defined in \cref{eqn:fedtpg_pred}.
    By utilizing an optimizer, e.g. SGD, we can estimate the unbiased gradient of $L_i(\theta^r; \mathcal{T}_i)$ with respect to $\theta^r$ and get the updated parameters $\theta^{r+1}_i$ after $K$ iterations with a learning rate $\eta^r$ as:
        \begin{align}\label{eqn:sgdk}
        \theta^{r+1}_i = \text{SGD}_K(\eta^r,\theta^r,\mathcal{T}_i,L_i)
     \end{align}
    \item \textbf{Step III:} After local few-shot training, all the remote clients in $\mathcal{S}^r$ send back their locally updated prompt generator $\theta^{r+1}_i $ to the server for aggregation:  $\theta^{r+1} = \frac{1}{|\mathcal{S}^r|} \sum_{i \in \mathcal{S}^r} \theta^{r+1}_{i}$.
\end{itemize}
After performing Step I-III for $R$ communication rounds, \gls*{fedtpg} obtains the final model parameters $\theta^{R}$. We argue that the proposed \gls*{fedtpg} can achieve the generalization goal from two aspects: (1) unlike existing prompt learning techniques that directly learn a fixed prompt vector, our TPG method captures a richer contextual and semantic information for each local classification task; (2) through the \gls*{fl} collaboration framework, diverse contextual and semantic information across multiple remote clients with different tasks 
benefit the model learning well.
Multiple clients encode text embeddings based on their distinct tasks, enabling the global model to serve a variety of contexts without overfitting to a specific task. Overall, the federated model can potentially learn a richer set of semantic features, and facilities better ``transfer learning'' capabilities, enabling the model to generalize well to both seen and new unseen tasks (that includes both classes and datasets). 



\section{Experiments}
\label{sec:experiment}

\begin{figure}[t]
\vspace{-20pt}
    \centering
	\begin{subfigure}[b]{0.32\linewidth}
		\includegraphics[width=\linewidth]{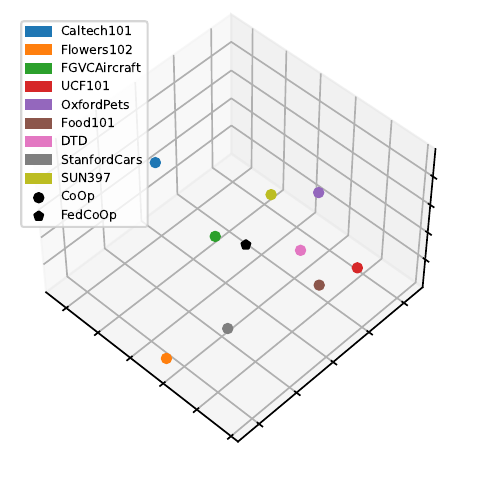}
	\caption{CoOp and FedCoOp's prompts}
	\end{subfigure}
	\begin{subfigure}[b]{0.32\linewidth}
		\includegraphics[width=\linewidth]{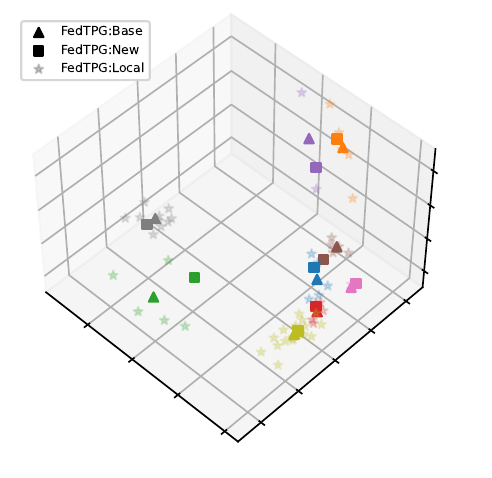}
    \caption{FedTPG's prompts}
	\end{subfigure}
 	\begin{subfigure}[b]{0.32\linewidth}
		\includegraphics[width=\linewidth]{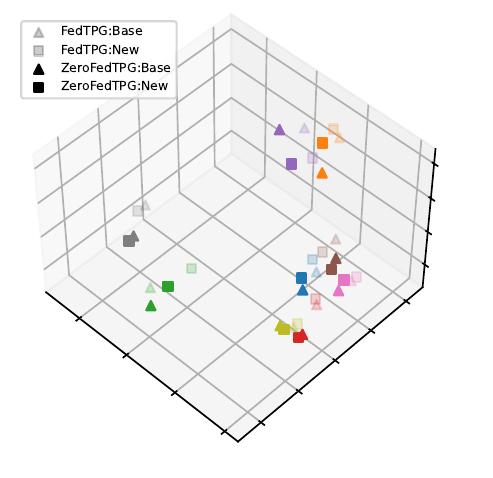}
    \caption{Zero-shot FedTPG's prompts}
	\end{subfigure}
    \caption{3D visualization (after PCA) of soft prompt vectors. (a) \gls*{coop} learns diverged prompt vectors on each dataset individually, while \gls*{fedcoop} learns one unified set of prompt vectors for tasks with various contexts (b) \gls*{fedtpg}'s prompt generator learned on bases classes generates context-aware prompt vectors for each task. (c) \gls*{fedtpg}'s prompt generator learned on ImageNet generates context-ware prompt vectors for nine unseen datasets aligned with the generated vectors in (b). }
    \label{fig:prompts}
\vspace{-10pt}
\end{figure}

We evaluate the proposed method \gls*{fedtpg} mainly on two benchmarks: (1) generalization to unseen related \textit{classes} in \Cref{sec:unseen_cls}, (2) generalization to unseen \textit{datasets} in \Cref{sec:unseen_data}. 
We also provide ablation studies to evaluate \gls*{fedtpg}'s robustness in various settings in \Cref{sec:ablation}. 
Below, we present our benchmark datasets, baselines, and implementation details.
\begin{table*}[t!]
    \vspace{-15pt}
    \caption{Accuracies ($\%$) on clients' local tasks (seen), base (seen) classes, and new (unseen) classes. \gls*{fedtpg} achieves the superior generalization performance over existing prompt learning methods and their \gls*{fl} variants, and the highest harmonic mean (HM) of three benchmark results.}
    \label{tab: base_new}
    \small
    \centering
    \subfloat[Average over 9 datasets.]{
    \begin{tabular}{c|ccc|c}
        \toprule
        & Local & Base & New & HM \\
        \midrule
CLIP&76.72& 70.52& 75.78&74.24\\
CoOp&\bf83.67
&71.49
&71.15
&75.01
\\
FedCoOp&81.75
&\bf74.50
&71.70
&75.75
\\
FedKgCoOp&78.38
&72.18
&75.87
&75.39
\\
\midrule
FedTPG&80.75&73.68&\bf76.02& \bf76.70 \\
        \bottomrule
    \end{tabular}}
\quad
   \subfloat[Caltech101.]{
    \begin{tabular}{c|ccc|c}
        \toprule
        & Local & Base & New & HM \\
        \midrule
CLIP&97.57&96.97&93.89&96.12\\
CoOp&
\bf97.79
&94.02
&93.1
&94.93
\\
FedCoOp&96.97&96.69&92.79&95.45\\
FedKgCoOp&97.65
&\bf97.24
&94.79
&96.54
\\
\midrule
FedTPG& 97.59 & 97.08 & \bf95.24 & \bf96.62\\
        \bottomrule
    \end{tabular}}\\
    \subfloat[Flowers102.]{
    \begin{tabular}{c|ccc|c}
        \toprule
        & Local & Base & New & HM \\
        \midrule
CLIP& 82.58&72.18&\bf77.94&77.33\\
CoOp&\bf97.27
&69.37
&71.95
&77.73\\
FedCoOp&94.44&\bf76.40&70.12&79.16\\
FedKgCoOp&84.59
&72.11
&77.06
&77.59
\\
\midrule
FedTPG& 90.76& 71.80& 77.76 &\bf79.35\\
        \bottomrule
    \end{tabular}}
\quad
   \subfloat[FGVCAircraft.]{
    \begin{tabular}{c|ccc|c}
        \toprule
        & Local & Base & New & HM \\
        \midrule
CLIP& 30.59&27.55&\bf35.81&30.96\\
CoOp&\bf36.88
&28.30
&28.59
&30.79
\\
FedCoOp&36.29&\bf32.41&30.95&33.07\\
FedKgCoOp&33.68
&29.79
&34.01
&32.37
\\
\midrule
FedTPG& 34.68& 30.82 & 35.18 & \bf33.44 \\
        \bottomrule
    \end{tabular}}\\
        \subfloat[UCF101.]{
    \begin{tabular}{c|ccc|c}
        \toprule
        & Local & Base & New & HM \\
        \midrule
CLIP&80.75&70.58&\bf77.5&76.04\\
CoOp&\bf88.37
&69.62
&68.09
&74.32
\\
FedCoOp&86.13&\bf75.65&70.60&76.93\\
FedKgCoOp&82.66
&73.14
&76.36
&77.19
\\
\midrule
FedTPG& 85.64 & 74.89 & 76.64 & \bf78.79\\
        \bottomrule
    \end{tabular}}
\quad
   \subfloat[OxfordPets.]{
    \begin{tabular}{c|ccc|c}
        \toprule
        & Local & Base & New & HM \\
        \midrule
CLIP& 91.33&91.33&\bf97.04&93.16\\
CoOp&\bf94.89
&\bf94.89
&96.60
&\bf95.46
\\
FedCoOp&93.31&93.32&95.39&94.00\\
FedKgCoOp&91.58
&91.58
&96.53
&93.17
\\
\midrule
FedTPG& 94.70& 94.69 & 95.79 & 95.06\\
        \bottomrule
    \end{tabular}}\\
        \subfloat[Food101.]{
    \begin{tabular}{c|ccc|c}
        \toprule
        & Local & Base & New & HM \\
        \midrule
CLIP&\bf94.39&90.16&91.25&91.90\\
CoOp&93.98
&88.20
&88.22
&90.05
\\
FedCoOp&93.52&88.63&88.47&90.15\\
FedKgCoOp&94.19
&\bf89.94
&\bf91.81
&\bf91.95
\\
\midrule
FedTPG& 94.09& 89.87 & 91.64 & 91.83\\
        \bottomrule
    \end{tabular}}
\quad
   \subfloat[DTD.]{
    \begin{tabular}{c|ccc|c}
        \toprule
        & Local & Base & New & HM \\
        \midrule
CLIP&53.13&53.01&58.21&54.68\\
CoOp&\bf72.34
&\bf72.34
&54.99
&\bf65.46
\\
FedCoOp&68.67&68.67&52.74&62.39\\
FedKgCoOp&58.76
&58.75
&59.61
&59.04\\
\midrule
FedTPG& 63.62& 63.62& \bf60.51 &62.55\\
        \bottomrule
    \end{tabular}}\\
        \subfloat[StanfordCars.]{
    \begin{tabular}{c|ccc|c}
        \toprule
        & Local & Base & New & HM \\
        \midrule
CLIP&71.51&63.44&74.9&69.61\\
CoOp&\bf78.65
&61.34
&70.17
&69.33
\\
FedCoOp&74.53&66.16&72.32&70.82\\
FedKgCoOp&71.89
&64.33
&\bf75.71
&70.32
\\
\midrule
FedTPG& 74.54& \bf66.34 & 74.26 &\bf71.50\\
        \bottomrule
    \end{tabular}}
\quad
   \subfloat[SUN397.]{
    \begin{tabular}{c|ccc|c}
        \toprule
        & Local & Base & New & HM \\
        \midrule
CLIP&88.66&69.41&75.46&77.05\\
CoOp&
\bf92.83
&65.29
&68.62
&73.78
\\
FedCoOp&91.93&72.34&71.89&77.70\\
FedKgCoOp&90.38
&72.72
&76.94
&79.34
\\
\midrule
FedTPG& 91.11& \bf74.01 & \bf77.13 &\bf80.10\\
        \bottomrule
    \end{tabular}}\\
        \vspace{-10pt}
\end{table*}

\textbf{Datasets} We employ nine image classification datasets that encompass a range of classification challenges. The benchmark includes Caltech101 \citep{fei2004learning} for generic objects classification; OxfordPets \citep{parkhi2012cats}, StanfordCars \citep{krause20133d}, Flowers102 \citep{nilsback2008automated}, Food101 \citep{bossard2014food} and FGVCAircraft \citep{maji2013fine} for classification on fine-grained categories; SUN397 \citep{xiao2010sun} for scene recognition; UCF101 \citep{soomro2012ucf101} for action recognition; DTD \citep{cimpoi2014describing} for texture classification. For evaluating domain generalization, we include ImageNetV2 \citep{recht2019imagenet}, ImageNet-Sketch \citep{wang2019learning}, ImageNet-A \citep{hendrycks2021natural}, and ImageNet-R \citep{hendrycks2021many}.


\textbf{Baselines.}
We compare \gls*{fedtpg} with 
\begin{enumerate*}[label=(\roman*)]
\item \gls*{clip} with hand-crafted text prompt template, e.g., ``a photo of a [class]'';
\item  \gls*{coop} \citep{zhou2022coop} with learnable prompt vectors replacing hand-crafted text prompts. CoOp is trained on each client individually to provide a baseline of local training.
\item  \gls*{fedcoop} \citep{guo2023promptfl}, a \gls*{fl} variant of CoOp. The unified prompt vectors are learned across multiple clients with federated averaging. 
\item  \gls*{fedkgcoop}, a \gls*{fl}-adapted variant of KgCoOp \citep{Yao_2023_CVPR} developed by us. We 
modify the original KgCoOp \citep{Yao_2023_CVPR} to the \gls*{fl} scheme as an additional baseline. KgCoOp improves over CoOp on generalization performance by adding a regularization of minimizing the discrepancy between the embeddings of learned prompts and the hand-crafted prompts. We develop \gls*{fedkgcoop} by combining KgCoOp with FedAvg \citep{mcmahan2017communication}.
\end{enumerate*}
For all \gls*{fl} methods, one unified model learned across clients is used for the evaluation of different datasets.  

\textbf{Implementation Details.}
All methods are built on a frozen CLIP with ViT-B/16 backbone.
Our proposed \gls*{fedtpg} learns a unified prompt generator parameterized by a four-head cross-attention layer with layer norm and a MLP ($h_\phi$) consisting of two linear layers with ReLU. The dimension of vectors $Q$, $K_{\mathcal{T}}$, $V_{\mathcal{T}}$ in the cross-attention layer, and linear layers in $h_\phi$ is 512. The length $m$ of generated prompt vectors is 4, and the dimension $d$ is 512. 
Similarly, all prompt learning-based baselines, including CoOp, FedCoOp, and FedKgCoOp, have also 4 learnable prompt vectors with a dimension of 512. Training
is done with SGD and an initial learning rate of 0.003, which is decayed by the cosine annealing rule. The number of communication rounds is 500. The batch size is 200. 

\subsection{Generalization to seen and unseen classes}
\label{sec:unseen_cls}
\textbf{Experimental setup.}
We split the classes of each dataset equally into two groups, one as base classes and the other as new classes.
Images from base classes are available for training, while the images from new classes are used for evaluating the generalization performance.
We consider a non-IID \gls*{fl} setting, where the base classes of all nine datasets are distributed to multiple clients.
Each client owns $n=20$ completely disjoint classes. We also consider a few-shot setting, where eight labeled images are available in each class for training. Note that all \gls{fl} methods learn one unified model or one unified set of prompt vectors on all clients jointly. 
We report the classification accuracies on clients' local classification tasks, on the base classes (combining classes from multiple clients), on the new classes in \Cref{tab: base_new}. We report the harmonic mean (HM) of these three accuracies showing the overall performance. 
All results are averaged over three independent runs.
 
\textbf{Quantitative results.}
As shown in \Cref{tab: base_new}(a), the proposed \gls*{fedtpg} achieves the best average accuracy on new classes, showing its advanced generalization ability. \gls*{fedtpg} also achieves second best performance on base classes and the highest harmonic mean which averages the accuracies on clients' local tasks, base classes, and new classes. Although the prompt generator is trained on local tasks consisting of a few classes, it generalizes well to a more complex classification task one the base classes (combining classes from multiple clients), and a novel classification task on the unseen classes. 
Due to the extreme non-IID setting, \gls*{coop} prompt vectors learned on each client dataset individually outperform the \gls*{fl} methods on the corresponding local task but fail to generalize to other base classes and new classes. Benefiting from learning across multiple clients, \gls*{fedcoop} improves over \gls*{coop} a lot on base classes. However, \gls*{fedcoop}'s performance gain is nearly zeroed out on new classes, highlighting the generalization challenge in federated prompt learning. 
Our newly-developed baseline \gls*{fedkgcoop} has an improved accuracy on new classes with a cost of performance degradation on base classes and local tasks, resulting from the difficulties of balancing the CLIP loss and the regularization term. 

\begin{table*}[tb]
    \vspace{-15pt}
    \caption{Accuracies ($\%$) on ImageNet (seen) and domain-shifted ImageNet variants (unseen). \gls*{fedtpg} consistently outperforms other baselines on both source dataset and domain-shifed datsets.}
\label{tab:domain}
    \small
    \centering
  
    \begin{tabular}{cccccc|c}
        \toprule
       &ImageNet & ImageNetV2 &ImageNet-S & ImageNet-A & ImageNet-R & Average\\
\cmidrule(r){1-2} \cmidrule(r){3-7}
CLIP&66.75 &60.79&	46.12&	47.79&	74.00&	57.18
 \\
FedCoOp&67.80 &61.59 &45.61	&48.78	&74.49	&	57.62
\\
FedKgCoOp&67.53 &61.60 &46.69&48.37	&74.71	& 57.84
\\
\cmidrule(r){1-2} \cmidrule(r){3-7}
FedTPG&\bf69.51&\bf62.90&\bf47.65&\bf49.97&\bf76.35&\bf59.22
\\
        \bottomrule  
    \end{tabular}

\end{table*}

\begin{table*}[t!]
    \caption{Accuracies ($\%$) on source (seen) and target (unseen) datasets. \gls*{fedtpg} consistently outperforms other federated prompt learning methods on both source dataset and unseen target datsets.}
\label{tab:zero_shot}
    \small
    \centering
    \resizebox{\linewidth}{!}{
    \begin{tabular}{ccccccccccc|c}
        \toprule
        &\multicolumn{1}{c}{Source} &\multicolumn{9}{c}{Target}\\
        \cmidrule(r){2-2} \cmidrule(r){3-12}
        &\rotatebox[origin=c]{60}{ImageNet} & \rotatebox[origin=c]{60}{Caltech101} & \rotatebox[origin=c]{60}{Flowers102} & \rotatebox[origin=c]{60}{FGVCAircraft} & \rotatebox[origin=c]{60}{UCF101} & \rotatebox[origin=c]{60}{OxfordPets} & \rotatebox[origin=c]{60}{Food101} & \rotatebox[origin=c]{60}{DTD} & \rotatebox[origin=c]{60}{StanfordCars} & \rotatebox[origin=c]{60}{SUN397}  & \rotatebox[origin=c]{60}{Average}\\
\cmidrule(r){1-2} \cmidrule(r){3-12}
CLIP&66.75& 92.90&\bf71.29&	\bf24.72&	66.75&	89.15&\bf	86.09&	44.33&\bf	65.29&	62.59&	67.01
\\
FedCoOp&67.80& 91.87&	68.13&	21.44&	64.13&	88.70&	85.85&	42.43&	63.59&	62.77&	65.43
\\
FedKgCoOp&67.53&93.63&	69.31&	23.06&	64.46&	88.55&	85.37&	44.74&	64.99&	63.85&	66.44
\\
\cmidrule(r){1-2} \cmidrule(r){3-12}
FedTPG&\bf69.51& \bf93.75 &70.04&23.22&	\bf64.72&	\bf90.60&	85.91&	\bf46.25&	63.98&	\bf66.78&	\bf67.25
\\
        \bottomrule
       
    \end{tabular}
    }
    \vspace{-10pt}
\end{table*}

\textbf{Qualitative analysis.}
We visualize the prompt vectors learned by \gls*{coop} on each dataset individually and the unified prompt vectors learned by \gls*{fedcoop} in \Cref{fig:prompts} (a), and the prompt vectors generated by \gls*{fedtpg} in \Cref{fig:prompts} (b). We can see that \gls*{coop} learns different optimal prompt vectors on each dataset.
However, the unified prompt vectors learned by \gls*{fedcoop} are not flexible enough to fit the context of all different datasets. In comparison, \gls*{fedtpg} learns to generate task-specific prompt vectors conditioning on the context-related text input. From \Cref{fig:prompts} (b) we can see the prompt vectors generated on clients (stars) sharing data from the same dataset are automatically clustered together, showcasing that the prompt generator learns to extract context information from the text input. Also, although the model is not trained on base-class classification and new-class classification, their associated generated prompt vectors (triangle for base, square for new) are clustered based on the dataset context accordingly, explaining \gls*{fedtpg}'s strong generalization ability. 


\subsection{Generalization to unseen datasets}
\label{sec:unseen_data}

\textbf{Experimental setup.}
For evaluating the generalization performance to unseen datasets, we train all models on ImageNet, and test the model on two benchmarks: (1) four variants of ImageNet containing various types of domain shifting: including ImageNetV2, ImageNet-Sketch, ImageNet-A, and ImageNet-R; (2) nine unseen datasets used in \Cref{tab: base_new}. Both are more challenging generalization problems since the task context can be completely different across datasets. We only consider \gls{fl} baselines in this setting.
We consider a non-IID setting with 200 clients. Each client owns $n=5$ completely disjoint classes. At each communication round, random $10\%$ clients contribute to the model update. We also consider a few-shot setting, where eight labeled images are available in each class. We report the accuracies on four variants of ImageNet in \Cref{tab:domain}, and the accuracies on nine unseen datasets in \Cref{tab:zero_shot}. All results are averaged over three independent runs.

\textbf{Results.}
Our proposed \gls*{fedtpg} improves over other federated prompt learning methods on ImageNet validation split and other variants of ImageNet consistently as shown in \Cref{tab:domain}. On the more challenging unseen datasets, \gls*{fedtpg} avoids the overfitting problem as the compared \gls*{fl} prompt methods, outperforming \gls*{clip} as shown in \Cref{tab:zero_shot}.  
Although the prompt generator is trained on images and class names from ImageNet, the model learns a generalizable function mapping the context-related text embeddings $\mathcal{T}$ to task-specific prompt vectors as visualized in \Cref{fig:prompts} (c), improving the classification accuracy on datasets with totally different context, e.g., from object recognition to texture classification. From \Cref{fig:prompts} (c) we can also see that the prompt vectors generated by the model trained on ImageNet are aligned with the prompt vectors generated by the model trained on these nine datasets, which demonstrates \gls*{fedtpg}'s cross-dataset transferability.


\subsection{Ablation studies}
\label{sec:ablation}
       

\begin{table*}[t!]
    \vspace{-15pt}
    \caption{Ablation study: three trials where each client owns $n=\{5,10,20\}$ disjoint classes. \gls*{fedtpg} consistently achieves the highest harmonic mean (HM) over \gls*{fedcoop} and \gls*{fedkgcoop}.  }
\label{tab:ablation}
    \small
    \centering
    \begin{tabular}{cccccccccc}
        \toprule
        & \multicolumn{3}{c}{n=5} & \multicolumn{3}{c}{n=10}  & \multicolumn{3}{c}{n=20}\\
        \cmidrule(r){2-4} \cmidrule(r){5-7} \cmidrule(r){8-10} 
        & Base & New& HM& Base & New& HM& Base & New& HM\\
    \cmidrule(r){2-4} \cmidrule(r){5-7} \cmidrule(r){8-10} 
FedCoOp& 69.53&70.05&69.69&\bf72.15&70.61&71.37 & \bf74.50& 71.70&73.07 \\
FedKgCoOp&70.83& \bf75.55&73.11&71.18&75.81 &73.42 & 72.18&75.87&73.98
\\
\midrule
FedTPG& \bf71.08& 75.51&\bf73.23&\bf72.15&\bf75.84& \bf73.95 & 73.68& \bf76.02& \bf74.83\\
        \bottomrule
       
    \end{tabular}
\end{table*}
\begin{figure}[t]
    \centering
	\begin{subfigure}[b]{0.42\linewidth}
		\includegraphics[width=\linewidth]{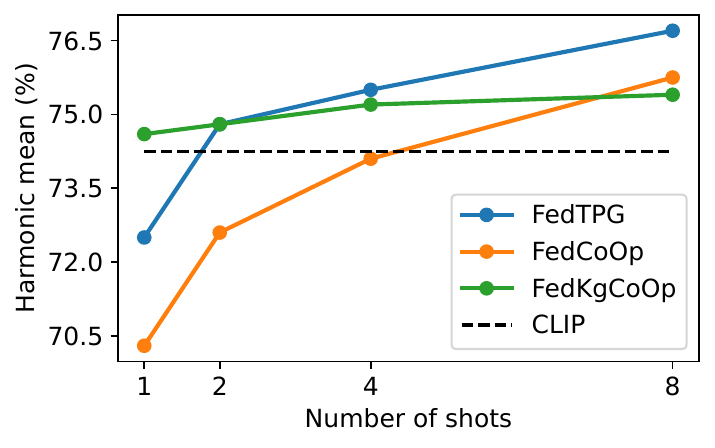}
	\caption{Number of shots}
	\end{subfigure}
	\begin{subfigure}[b]{0.42\linewidth}
		\includegraphics[width=\linewidth]{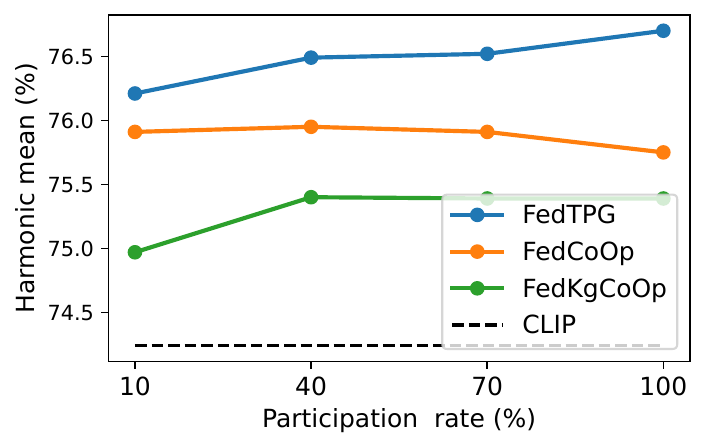}
    \caption{Participation rate of clients}
	\end{subfigure}
    \caption{(a) \gls*{fedtpg} gets improved when increasing the number of shots (for training), and has the best results when using more than one shot. (b) \gls*{fedtpg} is robust to the participation rate of clients.}
    \label{fig:ablation}
\vspace{-10pt}
\end{figure}
We provide three ablation studies to evaluate the robustness of \gls*{fedtpg} to the number of classes owned by each client, the number of shots, and the participation rate of clients in \gls*{fl}.

\textbf{Size of clients:} To understand the impact of the number of classes owned by the client, we conduct three trials where each client owns $n=\{5,10,20\}$ disjoint classes, and number of shots is 8. As shown in \Cref{tab:ablation}, \gls*{fedtpg} outperforms \gls*{fl} baselines in all cases in terms of the harmonic mean. 

\textbf{Number of shots:} We provide the model performance when using 1, 2, 4, 8 shots for training respectively in \Cref{fig:ablation}(a). We can see that \gls*{fedtpg} consistently improves over \gls*{fedcoop} in all cases, and outperforms \gls*{fedkgcoop} when the number of shots is larger than one.  

\textbf{Participation rate of clients:} We provide the model performance when the participation rate of clients varies from $10\%$ to $100\%$ in \Cref{fig:ablation}(b). We can see that \gls*{fedtpg} consistently outperforms \gls*{fedcoop} and \gls*{fedkgcoop} under varying participation rates of clients.
\section{Conclusion}
This paper addresses the fundamental challenge of generalization in adapting CLIP to the \gls{fl} setting. We propose a novel Federated Text-driven Prompt Generation (FedTPG) algorithm, which learns a unified prompt generator across multiple clients with various classification data collaboratively. The prompt generator learns to convert task-related text inputs to context-aware prompt vectors. Benefiting from context information in text inputs, the generated prompt vectors generalize well to unobserved classification problems. Our comprehensive experiments demonstrate FedTPG's superior generalization performance, outperforming existing \gls{fl} prompt methods by decent margins. 

\bibliography{refs}
\bibliographystyle{iclr2024_conference}
\newpage
\appendix
\section{Experiment Setup Details}

\subsection{Dataset and Hyper-parameter Details}
We follow the settings in \citet{zhou2022coop} to conduct the experiments in this paper with the nine classification datasets for generalization on seen to unseen classes, and four variants of ImageNet datasets for domain shifting, where the statistical details are presented in \Cref{tab:dataset}.

For each compared FL approach and each classification task, via grid search, the learning rate of the SGD optimizer was set to $\eta=0.003$ with a decay rate $1\mathrm{e}{-5}$ and a momentum of $0.9$. The local SGD training step is set to $K=1$. By default, all the experimental results in the paper are obtained by averaging from three runs with different random seeds.

\subsection{Federated Learning Setup Details}


\paragraph{Experimental Setup for Seen and Unseen Classes in \Cref{tab: base_new}}
To evaluate the generalization ability for the proposed FedTPG and compared FL approaches from in the paper, we monitor the model performance on the following three benchmark accuracies: (1) The local classification accuracy, representing the performance of local clients' classification tasks on local available classes; (2) The base classification accuracy, representing the performance against all seen classes (combining classes from multiple clients) in a dataset in the FL network;
(3) The new classification accuracy, which indicates the performance on unseen classes but within the domain of seen classes.
We report the harmonic mean (HM) of these three accuracies on each classification task, as shown in \Cref{tab: base_new}. 

In the FL data partition process for \Cref{tab: base_new}, we first split the classes of the considered 9 classification datasets equally into two groups $\mathcal{D}^{s}$ and $\mathcal{D}^{u}$, denotes seen and unseen groups respectively. Then we split the classes within $\mathcal{D}^{s}$ to the $30$ remote clients, where each remote client has $n = 20$ classes in each local dataset $\mathcal{D}_i$. For each class, the number of image-text paired data shots is set to 8. During the FL training process, the participation rate of remote clients is set to $100\%$ and the communication round is set to $500$.

\paragraph{Experimental Setup for Unseen Datasets in \Cref{tab:domain} and \Cref{tab:zero_shot}}
To evaluate the generalization ability of FedTPG on unseen datasets during training, we consider the following two settings: (1) Domain Shifting, where we monitor the performance of model by training with ImageNet and testing on four variants of ImageNet, including ImageNetV2, ImageNet-Sketch, ImageNet-A, and ImageNet-R; (2) Unseen Datasets, where we evaluate the performance of trained model in (1) on nine unseen datasets, including Caltech101, OxfordPets, StanfordCars, Flowers102, Food101, FGVCAircraft, SUN397, UCF101, and DTD. During the training process, we set the FL network with $200$ remote clients where each client has $n=5$ classes of $8$-shots training data disjointly. The participation rate of remote clients is set to $10\%$ that $|\mathcal{S}^r| = 20$ and the global communication round is set to $R=500$ to obtain $\theta^{R}$.

\paragraph{Experimental Setup for Ablation Study in \Cref{tab:ablation} and \Cref{fig:ablation}}
We study the impact of the number of classes owned by each client at \Cref{tab:ablation} from the introduced local, base and new classification accuracies with the same setup in \Cref{tab: base_new} where a full client participation is performed with $R=500$ and number of shots is $8$. Specifically, we perform the data partition with the disjoint rule during class splitting: when $n=5$, we set the number of clients to $119$; when $n=10$, we set the number of clients to $59$; and when $n=20$, we set the number of clients to $20$, respectively. 

The study of the number of shots is shown in \Cref{fig:ablation}(b), where we set the number of clients to $30$ with $n=20$ and the client participation rate is $100\%$ in each round where $R=500$. The study of the participation rate is shown in \Cref{fig:ablation}(b), where we set the number of clients to $30$ with $n=20$ and the number of shots is $8$.

Then, we monitor the impact of the FL client participation rate in each communication round as shown in \Cref{fig:ablation}(a). We formulate the FL network with 30 clients where $n=20$ and the number of shots is $8$. Four client participation rates in $\{10\%, 40\%, 70\%, 100\}\%$ are considered during the model training process with $R=500$.

\begin{table*}[tb]
    \vspace{-15pt}
    \caption{Dataset statistical details on class, training and test splits, prompt template.}
    \label{tab:dataset}
    \small
    \centering
    \begin{tabular}{c|cccc}
        \toprule
        Dataset & Classes & Train & Test& Hand-crafted prompt template \\
        \midrule
        ImageNet & 1000 & 1.28M & 50,000 & A photo of a \textbf{[class]} \\
        Caltech101 & 101 & 4,128 & 2,465& A photo of a \textbf{[class]} \\
        Flowers102 & 102 & 4,093 & 2,463 & A photo of a \textbf{[class]}, a type of flower \\
        FGVCAircraft & 100 & 3,334 & 3,333 & A photo of a \textbf{[class]}, a type of aircraft \\
        UCF101 & 101 & 7,639 & 3,783 & A photo of a person doing \textbf{[class]} \\
        OxfordPets & 37 & 2,944 & 3,369 & A photo of a \textbf{[class]}, a type of pet \\
        Food101 & 101 & 50,500 & 30,300 & A photo of a \textbf{[class]}, a type of food \\
        DTD & 47 & 2,820 & 1,692 & A photo of a \textbf{[class]}, a type of texture \\
        StanfordCars & 196 & 6,509 & 8,041 & A photo of a \textbf{[class]} \\
        SUN397 & 397 & 15,880 & 19,850 & A photo of a \textbf{[class]} \\
        \midrule
        ImageNetV2 & 1000 & N/A  & 10,000 & A photo of a \textbf{[class]} \\
        ImageNet-Sketch & 1000 & N/A & 50,889 & A photo of a \textbf{[class]} \\
        ImageNet-A & 200 & N/A  & 7500 & A photo of a \textbf{[class]} \\
        ImageNet-R & 200 & N/A  & 30,000 & A photo of a \textbf{[class]} \\
        \bottomrule
    \end{tabular}
\end{table*}

\section{Additional Results}
\Cref{tab:5} and \Cref{tab:10} show the detailed results of FedTPG and the compared FL baselines on the benchmark of seen and unseen classes with $n=5$ and $n=10$, respectively. The results of \Cref{tab:5} and \Cref{tab:10} are the detailed results of \Cref{tab:ablation} in the main paper, where we would like to claim that the HM results in the main paper are the harmonic mean of the base accuracy and the new accuracy, while the results in \Cref{tab:5} and \Cref{tab:10} are the harmonic mean of the local accuracy, the base accuracy and the new accuracy that leads to the difference in some columns. 

The results show that similar to the results of $n=20$ in \Cref{tab: base_new}, the proposed FedTPG achieves the best average accuracy on unseen classes, and achieves the best new performance for $3$ tasks while the second best new performance for most of the other tasks. We can also observe that as $n$ increases, the advantage of FedTPG against other approaches becomes more significant. This supports our theoretical claim that the unified prompt generator in FedTPG generalizes better on unobserved classification tasks, especially for challenging scenarios.  
\begin{table*}[t!]
    \caption{Accuracies ($\%$) on clients' local tasks (seen), base (seen) classes, and new (unseen) classes. Each client has labeled images from five disjoint classes. The number of shot is 8 and $n=5$.}
    \label{tab:5}
    \small
    \centering
    \subfloat[Average over 9 datasets.]{
    \begin{tabular}{c|ccc|c}
        \toprule
        & Local & Base & New & HM \\
        \midrule
CLIP&86.25&
70.52&
75.78&
76.98
\\
FedCoOp& 89.38&
69.53&
70.05&
74.74\\
FedKgCoOp&86.63&
70.83&
75.55&
77.12
\\
\midrule
FedTPG&87.78&
71.08&
75.51&
\bf77.51
 \\
        \bottomrule
    \end{tabular}}
\quad
   \subfloat[Caltech101.]{
    \begin{tabular}{c|ccc|c}
        \toprule
        & Local & Base & New & HM \\
        \midrule
CLIP&97.40&
96.97&
93.89&
96.06
\\
FedCoOp&97.19&
93.67&
92.14&
94.28
\\
FedKgCoOp&97.95&
96.57&
94.21&
\bf96.22
\\
\midrule
FedTPG& 97.31&
94.00&
94.43&
95.22\\
        \bottomrule
    \end{tabular}}\\
    \subfloat[Flowers102.]{
    \begin{tabular}{c|ccc|c}
        \toprule
        & Local & Base & New & HM \\
        \midrule
CLIP&91.12&
72.18&
77.94&
\bf79.66
\\
FedCoOp&97.89&
70.65&
74.47&
79.37
\\
FedKgCoOp&89.96&
70.27&
76.51&
78.09
\\
\midrule
FedTPG& 94.20&
70.23&
76.77&
79.20
\\
        \bottomrule
    \end{tabular}}
\quad
   \subfloat[FGVCAircraft.]{
    \begin{tabular}{c|ccc|c}
        \toprule
        & Local & Base & New & HM \\
        \midrule
CLIP&49.04&
27.55&
35.81&
35.45
\\
FedCoOp&55.82&
25.45&
26.57&
31.63
\\
FedKgCoOp&51.98&
28.89&
33.75&
\bf35.93
\\
\midrule
FedTPG&53.62&
26.38&
33.92&
34.87
 \\
        \bottomrule
    \end{tabular}}\\
        \subfloat[UCF101.]{
    \begin{tabular}{c|ccc|c}
        \toprule
        & Local & Base & New & HM \\
        \midrule
CLIP&88.78&
70.58&
77.50&
\bf78.25
\\
FedCoOp&90.71&
69.75&
65.33&
73.77
\\
FedKgCoOp&87.68&
70.06&
76.14&
77.29
\\
\midrule
FedTPG&88.53&
71.20&
75.96&
77.91
\\
        \bottomrule
    \end{tabular}}
\quad
   \subfloat[OxfordPets.]{
    \begin{tabular}{c|ccc|c}
        \toprule
        & Local & Base & New & HM \\
        \midrule
CLIP&96.75&
91.33&
97.04&
94.96
\\
FedCoOp&98.08&
91.92&
94.57&
94.79
\\
FedKgCoOp&96.65&
91.34&
96.16&
94.66
\\
\midrule
FedTPG&97.96&
91.39&
96.03&
\bf95.04
\\
        \bottomrule
    \end{tabular}}\\
        \subfloat[Foods102.]{
    \begin{tabular}{c|ccc|c}
        \toprule
        & Local & Base & New & HM \\
        \midrule
CLIP&97.57&
90.16&
91.25&
\bf92.88
\\
FedCoOp&97.17&
88.27&
86.67&
90.48
\\
FedKgCoOp&97.42&
89.59&
91.52&
92.72
\\
\midrule
FedTPG&97.34&
89.24&
91.31&
92.51
\\
        \bottomrule
    \end{tabular}}
\quad
   \subfloat[DTD.]{
    \begin{tabular}{c|ccc|c}
        \toprule
        & Local & Base & New & HM \\
        \midrule
CLIP&79.55&
53.01&
58.21&
61.71
\\
FedCoOp&86.94&
54.40&
51.45&
60.83
\\
FedKgCoOp&80.50&
55.47&
60.26&
63.77
\\
\midrule
FedTPG&82.72&
60.19&
61.53&
\bf66.73
\\
        \bottomrule
    \end{tabular}}\\
        \subfloat[StanfordCars.]{
    \begin{tabular}{c|ccc|c}
        \toprule
        & Local & Base & New & HM \\
        \midrule
CLIP&83.06&
63.44&
74.90&
72.90
\\
FedCoOp&86.06&
64.84&
71.77&
73.22
\\
FedKgCoOp&83.42&
63.84&
75.85&
\bf73.46
\\
\midrule
FedTPG& 83.75&
63.92&
72.35&
72.45
\\
        \bottomrule
    \end{tabular}}
\quad
   \subfloat[SUN397.]{
    \begin{tabular}{c|ccc|c}
        \toprule
        & Local & Base & New & HM \\
        \midrule
CLIP&93.02&
69.41&
75.46&
78.10
\\
FedCoOp&94.55&
66.83&
67.44&
74.32
\\
FedKgCoOp&94.12&
71.45&
75.52&
79.23
\\
\midrule
FedTPG&94.56&
73.17&
77.24&
\bf80.67
\\
        \bottomrule
    \end{tabular}}\\
\end{table*}

\begin{table*}[t!]
    \caption{Accuracies ($\%$) on clients' local tasks (seen), base (seen) classes, and new (unseen) classes. Each client has labeled images from ten disjoint classes. The number of shot is 8 and $n=10$.}
    \label{tab:10}
    \small
    \centering
    \subfloat[Average over 9 datasets.]{
    \begin{tabular}{c|ccc|c}
        \toprule
        & Local & Base & New & HM \\
        \midrule
CLIP&80.57& 70.52& 75.78&75.40\\
FedCoOp&85.64&72.15&70.61& 75.57\\
FedKgCoOp&81.39
&71.18
&75.81
&75.90\\
\midrule
FedTPG&83.49&72.17&75.84& \bf76.89 \\
        \bottomrule
    \end{tabular}}
\quad
   \subfloat[Caltech101.]{
    \begin{tabular}{c|ccc|c}
        \toprule
        & Local & Base & New & HM \\
        \midrule
CLIP&97.83&96.97&93.89&\bf96.20\\
FedCoOp&97.45&94.56&93.46& 95.13\\
FedKgCoOp&97.64
&96.80
&93.99
&96.12
\\
\midrule
FedTPG& 98.03 & 95.83 & 94.58 & 96.13\\
        \bottomrule
    \end{tabular}}\\
    \subfloat[Flowers102.]{
    \begin{tabular}{c|ccc|c}
        \toprule
        & Local & Base & New & HM \\
        \midrule
CLIP&84.58
&72.18
&77.94
&77.91
\\
FedCoOp&97.17&73.33&71.10& \bf78.96\\
FedKgCoOp&84.77
&71.93
&76.80
&77.48\\
\midrule
FedTPG& 90.03& 71.58& 77.08 &78.85\\
        \bottomrule
    \end{tabular}}
\quad
   \subfloat[FGVCAircraft.]{
    \begin{tabular}{c|ccc|c}
        \toprule
        & Local & Base & New & HM \\
        \midrule
CLIP&37.88
&27.55
&35.81
&33.10
\\
FedCoOp&44.00&27.23&25.76& 30.53\\
FedKgCoOp&38.53
&26.86
&35.06
&32.71\\
\midrule
FedTPG& 41.74& 28.44 & 35.05 & \bf34.21 \\
        \bottomrule
    \end{tabular}}\\
        \subfloat[UCF101.]{
    \begin{tabular}{c|ccc|c}
        \toprule
        & Local & Base & New & HM \\
        \midrule
CLIP&83.65
&70.58
&77.5
&76.87
\\
FedCoOp&87.56&73.53&71.76& 77.01\\
FedKgCoOp&84.00
&71.25
&76.11
&76.77\\
\midrule
FedTPG& 85.78 & 72.15 & 76.05 & \bf77.59\\
        \bottomrule
    \end{tabular}}
\quad
   \subfloat[OxfordPets.]{
    \begin{tabular}{c|ccc|c}
        \toprule
        & Local & Base & New & HM \\
        \midrule
CLIP&93.26
&91.33
&97.04
&93.82
 \\
FedCoOp&95.95&92.36&91.60& 93.27\\
FedKgCoOp&92.55
&90.32
&96.36
&93.01\\
\midrule
FedTPG& 95.86& 93.92 & 96.73 & \bf95.48\\
        \bottomrule
    \end{tabular}}\\
        \subfloat[Foods102.]{
    \begin{tabular}{c|ccc|c}
        \toprule
        & Local & Base & New & HM \\
        \midrule
CLIP&95.94
&90.16
&91.25
&\bf92.38
\\
FedCoOp&95.18&88.21&89.91& 90.72\\
FedKgCoOp&95.81
&89.88
&91.66
&\bf92.38\\
\midrule
FedTPG& 95.73& 89.93 & 91.63 & 92.36\\
        \bottomrule
    \end{tabular}}
\quad
   \subfloat[DTD.]{
    \begin{tabular}{c|ccc|c}
        \toprule
        & Local & Base & New & HM \\
        \midrule
CLIP&62.74
&53.01
&58.21
&57.71
\\
FedCoOp&78.15&63.11&49.65& 61.50\\
FedKgCoOp&68.10
&57.12
&60.26
&61.49\\
\midrule
FedTPG& 71.41& 59.52& 60.18 &\bf63.26\\
        \bottomrule
    \end{tabular}}\\
        \subfloat[StanfordCars.]{
    \begin{tabular}{c|ccc|c}
        \toprule
        & Local & Base & New & HM \\
        \midrule
CLIP&78.29
&63.44
&74.9
&71.62
\\
FedCoOp&81.23&65.76&70.93& 72.09\\
FedKgCoOp&78.82
&64.13
&75.52
&72.25\\
\midrule
FedTPG& 80.15& 65.33 & 74.62 &\bf72.84\\
        \bottomrule
    \end{tabular}}
\quad
   \subfloat[SUN397.]{
    \begin{tabular}{c|ccc|c}
        \toprule
        & Local & Base & New & HM \\
        \midrule
CLIP&90.96
&69.41
&75.46
&77.61
\\
FedCoOp&94.07&71.32&72.10& 77.88\\
FedKgCoOp&92.28
&72.36
&76.47
&79.51\\
\midrule
FedTPG& 92.71& 72.90 & 76.62 &\bf79.88\\
        \bottomrule
    \end{tabular}}\\
\end{table*}

\end{document}